# When Auditors Fabricate: Batch-Size Degradation and Confident Hallucination in LLM Detection of Planted Document Contamination

Karan Parekh, Sanjana Pendyala Ravinder, Sana Mhapsekar, Medina Maloku

University of North Texas

---

## Abstract

Large language models are increasingly proposed as automated auditors of document quality, yet their reliability as detectors of planted errors is poorly characterised. We construct a contaminated corpus of 150 academic papers spanning supply chain management and medical research, injecting 450 known contaminants of three types: typographical corruption, semantic reversal, and absurd out-of-context insertion. We then evaluate Google Gemini 3.0 Pro's ability to recover a 180-contaminant answer-key subset across 60 documents under three prompting regimes of increasing scale: single document, small batch, and large batch. Detection is unreliable even at small scale and collapses entirely at large scale: 50% recovery on single documents and 60% on small batches, a difference this sample cannot resolve, against 2.8% on large batches. The failure mode at scale is not abstention but fabrication. Rather than reporting incomplete processing, the model produced confident findings including invented contaminants of its own, absurdities such as “telepathic squirrel” and “quantum-powered toaster” that mimic the style of the planted material but do not appear in any document. Detection also varies by contamination type: absurd insertions were recovered at 75% in completed evaluations, while semantic reversals and typographical corruptions were each recovered at only 50%. The corruptions most likely to occur in the wild, plausible ones, are the ones most often missed. We conclude that LLM document auditing degrades not gracefully but deceptively, and outline the harness such systems require: bounded batch sizes, direct content injection, and mechanical verification of every reported finding against source text.

## 1. Introduction

If an LLM is asked to audit more documents than it can process, what does it do? The safe behaviour is abstention, and models can in principle recognise the limits of their own knowledge and be tuned to decline (Kadavath et al., 2022; Zhang et al., 2024). The behaviour we observe is fabrication, and fabrication in the expected genre: asked to find planted absurdities, the model invented plausible-sounding absurdities of its own.

This matters because LLMs are being deployed as quality gates: reviewing documents, checking claims, flagging errors (Zheng et al., 2023; Liang et al., 2024; Lovering et al., 2025; Son et al., 2025). Such deployments implicitly assume the auditor is at worst incomplete, not actively misleading. Hallucination is by now well documented as a general property of language generation (Ji et al., 2023; Huang et al., 2025), but the auditing setting inverts the usual concern: here the model's output is itself the quality signal, so a fabricated finding does not merely mislead a reader, it corrupts the control that was supposed to catch the error. We test that assumption by planting known contaminants in real academic documents and measuring what a model reports back as workload increases.

A note on terminology. In the LLM literature, contamination usually refers to benchmark or training data leakage, the presence of evaluation material in a model's pretraining corpus (Magar and Schwartz, 2022; Sainz et al., 2023; Golchin and Surdeanu, 2024). We use the word in a narrower and unrelated sense: deliberately planted defects inside the documents a model is asked to audit.

Contributions:

1. A contaminated-corpus construction method with a three-type taxonomy spanning surface corruption (typos), semantic corruption (meaning reversal), and contextual corruption (absurd insertion), applied to 150 real academic PDFs (450 contaminants).
2. A scaled evaluation protocol measuring recovery of a 180-contaminant answer key across 60 documents at three batch sizes.
3. Documentation of a deceptive failure mode: under load the model does not report failure, it generates fluent, well-structured, wholly fabricated audit findings, including invented contaminants styled after the expected ones.

## 2. Corpus and Contamination Method

**Source documents.** 150 published academic papers in two domains, supply chain management and medical and pharmaceutical sciences, chosen to test generalisation across technical vocabularies.

**Contamination taxonomy.** Each contaminated document received three planted contaminants, one per type:

| Type | Mechanism | Examples (actual) | Detectability hypothesis |
|---|---|---|---|
| Typo | Surface corruption of a real word | "efficency", "buisness", "knowlege" | Detectable by spelling alone |
| Conflicting | Semantic reversal of a directional claim | "improvements" becomes "deteriorations"; "robust" becomes "fragile" | Requires understanding the claim |
| Nonsense | Absurd out-of-context insertion | "supply chain" becomes "dolphin choir"; "parameter" becomes "parachute" | Contextually obvious |

Every contaminant was logged with contaminant ID, document ID, page, and original and replacement text, forming a complete answer key.

**Deployment.** The corpus was published as a reference-accessible knowledge base, enabling evaluation via document references rather than inline content.

## 3. Evaluation Protocol

Google Gemini 3.0 Pro (free tier) was prompted to identify contaminants under three regimes:

- **Part I, single document.** One supply chain and one medical document. 6 contaminants.
- **Part II, small batch.** One batch of 7 supply chain documents and one of 3 medical documents. 30 contaminants.
- **Part III, large batch.** 48 documents in six batches of 8. 144 contaminants.

Responses were scored programmatically against the answer key (exact and fuzzy matching with manual adjudication of page and type mismatches).

## 4. Results

### 4.1 Detection collapses with batch size

| Regime | Docs | Contaminants | Recovered | Rate |
| --- | --- | --- | --- | --- |
| Part I, single document | 2 | 6 | 3 | 50% |
| Part II, small batch | 10 | 30 | 18 | 60% |
| Part III, large batch | 48 | 144 | 4 | 2.8% |
| **Overall** | **60** | **180** | **25** | **13.9%** |

Reported as 95% Wilson intervals, Part I is 18.8% to 81.2%, Part II is 42.3% to 75.4%, and Part III is 1.1% to 6.9%. Parts I and II are therefore not distinguishable in this sample (two-proportion $z = 0.45$, $p = 0.65$), and only the collapse from small to large batch is a measured effect ($z = 8.58$, p below 1e-17). We accordingly claim one effect rather than a graded curve, and note that Part I rests on two documents. Within Part II the supply chain batch of 7 recovered 15/21 (71.4%) while the medical batch of 3 recovered only 3/9 (33.3%), indicating domain and document effects alongside the batch-size effect. The single supply chain document in Part I was an outlier at 0/3 while its medical counterpart scored 3/3. In Part III, four of six batches recovered zero contaminants; no batch recovered more than two of twenty-four. Position and length effects in long-context models are well established. Accuracy falls when the relevant material sits in the middle of a long context rather than at either end (Liu et al., 2024), and models that score near perfectly on simple needle-in-a-haystack probes frequently fail more demanding tasks well below their advertised context length (Hsieh et al., 2024). What our result adds is the shape of the failure at the point of overload: the model does not return a lower score on the planted items, it returns a different set of items.

### 4.2 The failure mode at scale is confident fabrication

In the large-batch regime the model returned findings for the requested documents in fluent, well-structured form. The findings were fabricated. The scoring log records the model reporting invented contaminants including “quantum-powered toaster”, “telepathic squirrel”, “disco-dancing warehouse”, “flying pancake”, “magic carpet”, “pizza cutter”, and “pogo stick equation”. None of these strings appear in any corpus document. The actual planted contaminants in those same documents, items such as “dolphin choir”, “efficency” and “deteriorations”, went almost entirely unreported (4 of 144 recovered).

Two properties of this failure mode deserve emphasis. First, it is **genre-consistent**: the invented contaminants imitate the absurdist style of the real planted ones, suggesting the model inferred the kind of answer expected and generated to that expectation rather than to the documents. A related pattern is documented for sycophancy, where models produce what a reader is expected to prefer rather than what the evidence supports, in part because preference data rewards it (Sharma et al., 2024). Second, it is **formally indistinguishable from success**. The fabricated audits are fluent, correctly structured, and returned in exactly the requested table format, with page numbers, putative original text and reasoning supplied for every invented item. Nothing in their form separates them from the genuine findings of Parts I and II. This is what makes the failure dangerous rather than merely inconvenient, because fluency and confidence are poor proxies for correctness. LLM judges can be swayed by ordering and presentation rather than substance (Wang et al., 2024), verbalised confidence is systematically overstated (Kadavath et al., 2022; Xiong et al., 2024), and human raters and preference models reward convincingly written but incorrect answers (Sharma et al., 2024). A downstream consumer of these audits has no available signal that separates the fabricated run from the successful one. Low recall on error finding is not unique to our setting either: across 83 published papers containing errors severe enough to have prompted errata or retraction, no frontier model exceeded 21.1% recall or 6.1% precision, with poorly calibrated confidence and low run-to-run consistency (Son et al., 2025).

One partial recovery illustrates the boundary: for a document contaminated with “volcano dance”, the model reported “disco dance”, the right location and genre with the wrong content, consistent with reconstruction from partial context rather than reading.

### 4.3 Detection asymmetry by contamination type

Within completed evaluations (Parts I and II, 36 contaminants):

| Type | Recovered | Rate |
|---|---|---|
| Nonsense (absurd insertion) | 9/12 | 75% |
| Typo (surface corruption) | 6/12 | 50% |
| Conflicting (semantic reversal) | 6/12 | 50% |

One property of the corpus must be disclosed before these rates are read. The absurd insertion was not varied uniformly: the string “dolphin choir” accounts for 37 of the 60 nonsense contaminants in the scored set, so within a batch these items are not independent. Recomputing the completed evaluations with the repeated string separated leaves the rate unchanged, 6/8 for “dolphin choir” against 3/4 for the varied absurdities, but the varied subset is too small to be reassuring and a replication should vary the insertion by document. Contextually absurd insertions were caught most reliably. Semantic reversals, which preserve grammatical plausibility while inverting a claim’s direction, and simple typographical corruptions were each missed half the time. **The contaminations most representative of real-world corruption, plausible ones, are precisely the ones the model misses most.**

Across all 180 contaminants including the fabrication-dominated large batches, recovery was 18.3% for nonsense, 13.3% for conflicting, and 10% for typos. Independent work reports a similar order of magnitude: on expert-inserted inconsistencies in long technical documents, the strongest model tested recovered 64% of them and every model tested missed roughly half (Lovering et al., 2025).

### 4.4 False positives

In the single-document supply chain trial the model reported six contaminants, none planted. Whether these are hallucinations or genuine pre-existing defects in the source paper is unresolved; either way, an auditor whose findings require their own audit loses much of its value. The ambiguity is not ours alone. Lovering et al. (2025) report that of the unplanted items their models flagged, a large majority were judged on review to be genuine pre-existing errors in the source papers, which means unplanted flags cannot be treated as noise without adjudication. Settling this requires expert review of the source document rather than answer-key matching, and we did not perform it.

## 5. Discussion and Deployment Implications

1. **Bound the batch.** Detection was serviceable at 1 to 10 documents and collapsed at 48. Auditing pipelines should shard aggressively regardless of nominal context capacity.
2. **Treat the access method as an open confound.** Every regime here supplied documents by URL and asked the model to fetch them, so batch size and retrieval load vary together and this study cannot separate context overload from silent retrieval failure. Both mechanisms predict what we observed. This is the principal confound of the design and the first experiment a replication should run: hold batch size fixed and vary only whether document text is referenced or placed directly in context.
3. **Verify mechanically.** Every reported finding should be string-checked against source text before acceptance. Our scoring pipeline did this by construction; the fabricated findings would have been rejected automatically. Production systems typically lack this loop. Decomposing a generation into checkable units and verifying each against a source is an established mitigation, whether by atomic fact scoring against a knowledge source (Min et al., 2023) or by planning verification questions and answering them independently of the draft (Dhuliawala et al., 2024). In the auditing setting the check is cheaper still, because a reported finding is a claim that a specific string appears in a specific document, which is decidable by exact match.
4. **Design for loud failure.** The model never reported inability to process the large batches. Wrapping systems must detect and surface incompleteness themselves, because the model will not volunteer it, and its silence is dressed as an answer. Refusal-aware tuning shows that declining to answer can be taught as a general skill (Zhang et al., 2024), but an integrator cannot assume it is present in a model they do not control.

## 6. Limitations

Single model (Gemini 3.0 Pro, free tier), single scored run per regime, no prompt-variation or temperature sweep. The answer key covers 180 of the 450 planted contaminants; the remainder were not evaluated. Contaminant density was fixed at three per document, one per type. The 60 scored documents were drawn by a seeded random shuffle stratified by domain, so the 180 scored contaminants are a random subset of the 450 planted rather than a selected one. The absurd insertion was not varied uniformly across documents, as noted in section 4.3. Most importantly, because documents were referenced by URL in every regime, batch size is confounded with retrieval load, and the mechanism behind the collapse is not established by this study. Results reflect model access at evaluation time and may not transfer to current models; the protocol is model-agnostic and inexpensive to repeat.

## 7. Conclusion

Planted-contamination evaluation is a cheap, repeatable way to measure whether an LLM auditor can be trusted. Ours cannot be, unsupervised: detection is unreliable even on single documents and collapses at batch scale, the collapse presents as confident output rather than surfaced failure, the model invents findings in the genre it expects, and the most plausible corruptions are the least detected. None of this argues against LLM document auditing. It is the specification for the harness such auditing must run inside.

---

## Data and code availability

Evaluation pipeline, prompts, scoring scripts and full detection logs: github.com/karanparekh14/llm-contamination-detection-eval. Contaminated corpus and answer key: github.com/karanparekh14/genai-rag-qa-portfolio.

## Acknowledgements

This work originated in coursework for ADTA-DAST 5770 at the University of North Texas.